\documentclass[10pt,twocolumn,letterpaper]{article}

\usepackage{iccv}
\usepackage{booktabs}
\usepackage{times}
\usepackage{epsfig}
\usepackage{graphicx}
\usepackage{amsmath}
\usepackage{amssymb}
\usepackage{bbm}
\usepackage[font=small,skip=0pt,labelsep=period]{caption}
\usepackage{subcaption}
\usepackage{multirow}
\usepackage{eqparbox,array}
\usepackage{makecell}

\usepackage{authblk}
\makeatletter
\renewcommand\AB@affilsepx{, \protect\Affilfont}
\makeatother
\usepackage{lipsum}
\newcommand\blfootnote[1]{%
\begingroup
\renewcommand\thefootnote{}\footnote{#1}%
\addtocounter{footnote}{-1}%
\endgroup
}

\usepackage[pagebackref=true,breaklinks=true,letterpaper=true,colorlinks,bookmarks=false]{hyperref}

\iccvfinalcopy 

\def\iccvPaperID{8338} 

\ificcvfinal\fi

\begin{document}

\title{Product1M: Towards Weakly Supervised Instance-Level Product Retrieval \\ via Cross-Modal Pretraining}
\author[1$\dagger$]{Xunlin Zhan}
\author[1$\dagger$]{Yangxin Wu}
\author[1]{Xiao Dong}
\author[2]{Yunchao Wei}
\author[3]{Minlong Lu}
\author[3]{Yichi Zhang}
\author[4]{Hang Xu}
\author[1$\star$]{Xiaodan Liang}
\affil[1]{Sun Yat-sen University}
\affil[2]{Beijing Jiaotong University}
\affil[3]{Alibaba Group}
\affil[4]{Huawei Noah’s Ark Lab
\protect\\
\textit {\small  \{zhanxlin, wuyx29\}@mail2.sysu.edu.cn, \{dx.icandoit,wychao1987,chromexbjxh,xdliang328\}@gmail.com, ymlml@zju.edu.cn, yichi.zyc@alibaba-inc.com}}

\maketitle
\ificcvfinal\thispagestyle{empty}\fi

\begin{abstract}
Nowadays, customer's demands for E-commerce are more diversified, which introduces more complications to the product retrieval industry.
Previous methods are either subject to single-modal input or perform supervised image-level product retrieval, thus fail to accommodate real-life scenarios where enormous weakly annotated multi-modal data are present.
In this paper, we investigate a more realistic setting that aims to perform weakly-supervised multi-modal instance-level product retrieval among fine-grained product categories.
To promote the study of this challenging task, we contribute Product1M, one of the largest multi-modal cosmetic datasets for real-world instance-level retrieval.
Notably, Product1M contains over 1 million image-caption pairs and consists of two sample types, i.e., single-product and multi-product samples, which encompass a wide variety of cosmetics brands.
In addition to the great diversity, Product1M enjoys several appealing characteristics including fine-grained categories, complex combinations, and fuzzy correspondence that well mimic the real-world scenes.
Moreover, we propose a novel model named Cross-modal contrAstive Product Transformer for instance-level prodUct REtrieval (CAPTURE), that excels in capturing the potential synergy between multi-modal inputs via a hybrid-stream transformer in a self-supervised manner.
CAPTURE generates discriminative instance features via masked multi-modal learning as well as cross-modal contrastive pretraining and it outperforms several SOTA cross-modal baselines.
Extensive ablation studies well demonstrate the effectiveness and the generalization capacity of our model.
Dataset and codes are available at \url{https://github.com/zhanxlin/Product1M}.
\blfootnote{$\dagger$ Equal contribution. $\star$ Corresponding Author.}
\end{abstract}

\section{Introduction}

\begin{figure}[h]
\setlength{\belowcaptionskip}{-0.6cm}
    \centering
    \includegraphics[width=0.5\textwidth]{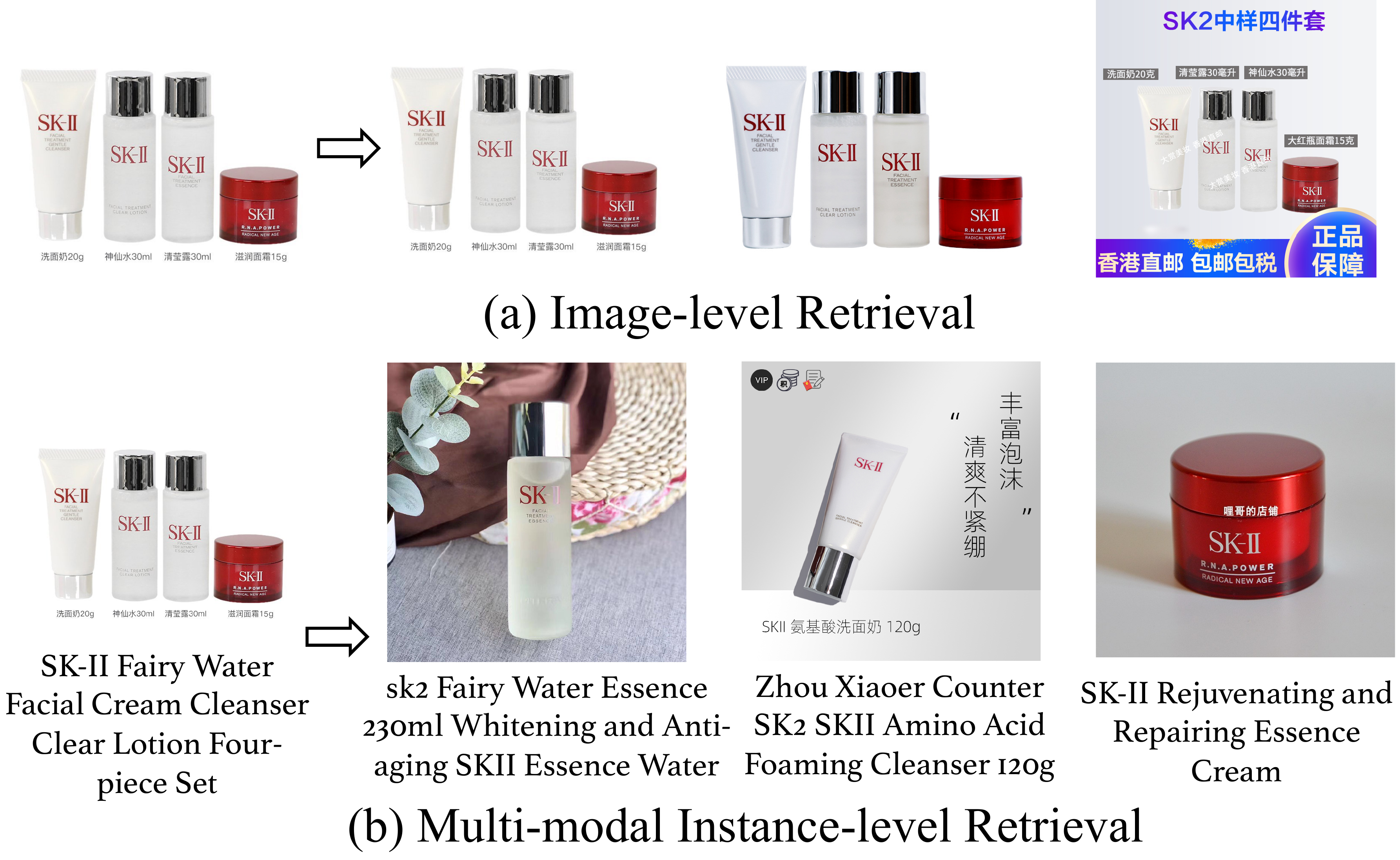}
    \caption{
    Our proposed task performs \textbf{instance-level} retrieval among \textbf{multi-modal} data.}
    \label{fig_task_difference}
\end{figure}
The past two decades have witnessed the high enrichment of the commodity types and the diversification of online customer's demand in E-commerce.
On the one hand, online merchandise has increasingly diversified categories and a large proportion of them are exhibited as a product portfolio where multiple instances of different products exist in one image.
On the other hand, online customers or merchants may want to retrieve the single product in a portfolio for price comparison \cite{vogler2016cancer} or online commodity recommendation~\cite{rahayu2017systematic}.
Furthermore, with the ever-accelerating accumulation of heterogeneous data generated by multimedia, it remains a problem how an algorithm can handle large-scale and weakly annotated data \cite{wang2016comprehensive} to perform multi-modal retrieval.

\begin{table*}
\centering
\setlength{\belowcaptionskip}{-0.5cm}
\begin{tabular}{c|cccc|ccc}
\toprule[1pt]
{{Dataset}}& {\footnotesize{}{\#samples}} &  {\footnotesize{}{\#categories}} &  {\footnotesize{}{\#instances}}
&  {\footnotesize{}{\#obj/img}}
&  {\footnotesize{}{weak supervision}}
&  {\footnotesize{}{multi-modal}}
&  {\footnotesize{}{instance-level retrieval}}
\tabularnewline
\midrule[1pt]
RPC checkout \cite{wei2019rpc} & 30,000 & 200 & 367,935 & 12.26 &  & & \tabularnewline
Twitter100k \cite{hu2017twitter100k} & 100,000 & - & - & - & \checkmark &\checkmark & \tabularnewline
INRIA-Websearch \cite{Krapac2010ImprovingWI} &  71,478 & 353  & - & - &\checkmark &\checkmark & \tabularnewline
Dress Retrieval \cite{corbiere2017leveraging} & 20,200 & - & $\sim$20,200 & $\sim$1.0&\checkmark &\checkmark & \tabularnewline
\textbf{Product1M(Ours)} & \textbf{1,182,083}  & \textbf{458} & \textit{92,200} & \textit{2.83} & \checkmark & \checkmark  & \checkmark \tabularnewline

\bottomrule[1pt]
\end{tabular}
\caption{Comparisons between different datasets. `-' indicates inapplicable. The \#instances and \#obj/img of Product1M are in italics since there are no instance labels for the \textit{train} set and we only count the instances in the \textit{val} and \textit{test} set. Product1M is one of the largest multi-modal datasets as well as the first dataset specifically tailored for real-world instance-level retrieval scenarios.}
\label{tab_statistics}
\end{table*}

In this paper, we explore a realistic problem: \textit{how to perform instance-level\footnote{Instance-level product retrieval refers to the retrieval of all single products existed in a product portfolio image.} fine-grained product retrieval given the large-scale weakly annotated multi-modal data?}
We compare different paradigms of retrieval in Figure \ref{fig_task_difference}.
As can be seen, image-level retrieval tends to return trivial results since it does not distinguish different instances, while multi-modal instance-level retrieval is more favorable for searching for various kinds of products among multi-modal data.
Despite the generality and the practical value of this problem, it is not well studied due to the lack of real-world datasets and a clear problem definition.
In the literature of product retrieval, intra-modal \cite{qi2016sketch,bai2018optimization,qayyum2017medical,nurmi2008product} and cross-modal retrieval \cite{wang2017adversarial,feng2014cross,wei2016cross,cao2016deep,wang2015joint,deng2018triplet} take as input single-modal information, e.g., an image or a piece of text, and performs matching search between separate data points.
Unfortunately, such retrieval schemes significantly restrict their use in many scenarios where multi-modal information exists in both the queries and targets.
More importantly, previous works focus on the relatively simple case, i.e., image-level \footnote{Image-level product retrieval refers to recognizing a specific product instance in a single-product image.} retrieval for single-product images \cite{kuang2019fashion,gu2018multi} and the instance-level nature of retrieval is unexplored.

To bridge this gap and advance the related research, we collect a large-scale dataset, named Product1M, proposed for multi-modal instance-level retrieval.
Product1M contains over 1 million image-caption pairs and consists of two types of samples, i.e., single-product and multi-product samples.
Each single-product sample belongs to a fine-grained category and the inter-category difference is subtle.
The multi-product samples are of great diversity, resulting in complex combinations and fuzzy correspondence that well mimic the real-world scenarios.
To the best of our knowledge, Product1M is one of the largest multi-modal datasets as well as the first dataset specifically tailored for real-world  multi-modal instance-level retrieval scenarios.

In addition to the constructed dataset, we also propose a novel self-supervised training framework that extracts representative instance-level features from large-scale weakly annotated data.
Specifically, we first train a multi-product detector from pseudo-labels by incorporating a simple yet effective data augmentation scheme.
Then, CAPTURE is proposed to capture the potential synergy of images and texts via several pretext tasks.
We showcase that some prevailing cross-modal pretraining methods \cite{lu2019vilbert,li2019visualbert,chen2020uniter,tan2019lxmert} might be flawed under the multi-instance setting due to the design defects in the network architecture or the inappropriate pretext task.
In contrast, CAPTURE utilizes a hybrid-stream architecture that encodes data of different modalities separately and fuses them in a unified way, which is experimentally shown to be beneficial for our proposed task.
Moreover, we introduce the cross-modal contrastive loss to enforce CAPTURE to reach alignment between image and texts, which avoids the mismatch issue incurred by the inappropriate pretext task.

Crucially, CAPTURE surpasses the SOTA cross-modal baselines in terms of all main metrics by a large margin.
We further conduct extensive ablation experiments to demonstrate the generalization capacity of CAPTURE and explore several critical factors of our proposed task.
We hope the proposed Product1M, CAPTURE, and solid baselines can help advance future research on real-world retrieval.

\begin{figure*}
\setlength{\belowcaptionskip}{-0.6cm}
    \centering
    \includegraphics[width=1.0\textwidth]{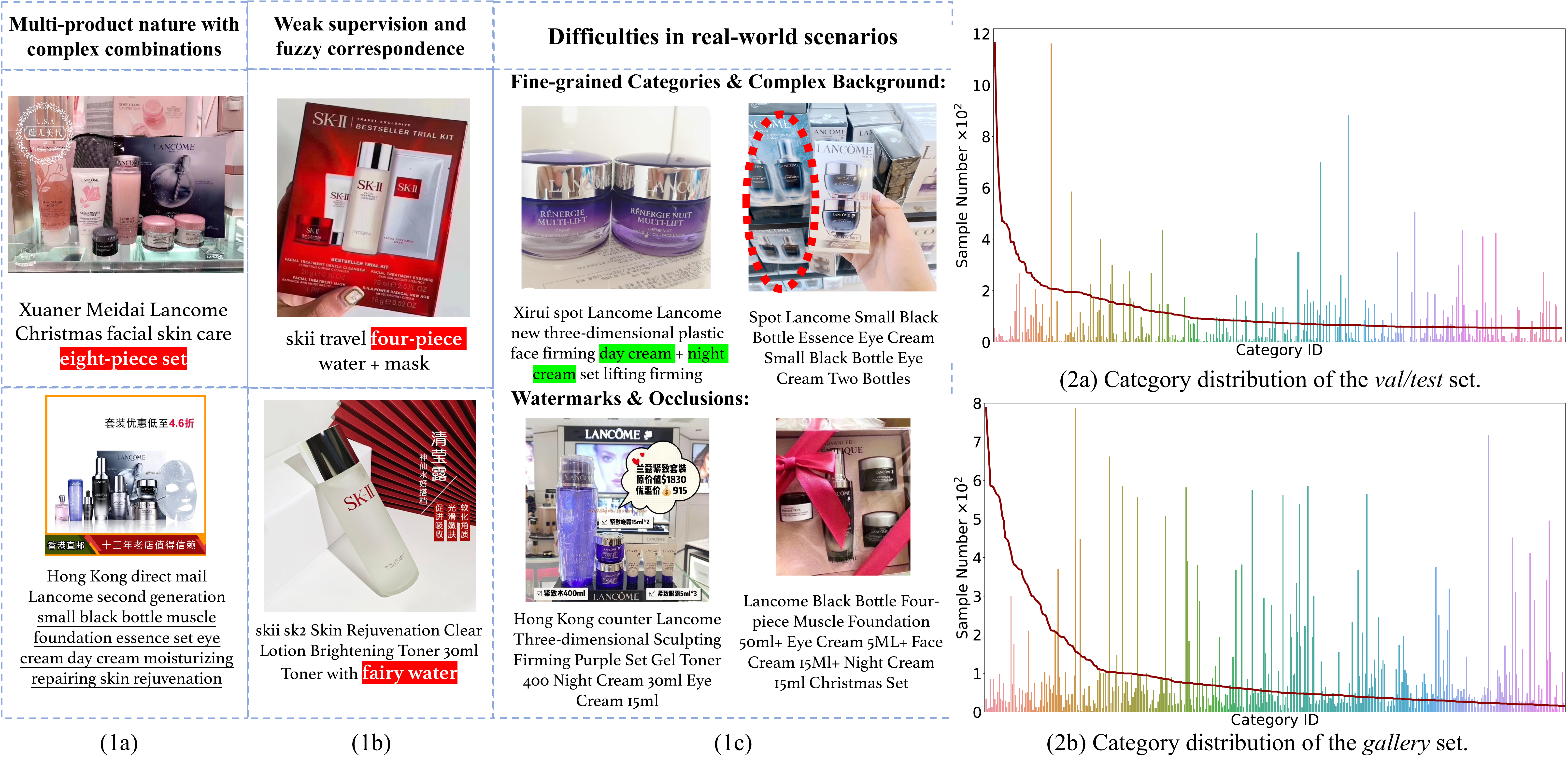}
    \caption{Characteristics and statistics of Product1M: (1a) Complex combinations of single-product; (1b) Weak supervision and fuzzy correspondence; (1c) Difficulties in real-world scenarios;
    (2) Long-tailed category distribution of Product1M. The line displays the sample number of each category in decreasing order. Product1M contains a wide variety of categories and the long-tailed class distribution aligns well with real-world scenarios.}
    \label{fig_challenge_statistics}
\end{figure*}

\section{Related Work}


\noindent\textbf{Intra- and Cross-Modal Retrieval.} Intra-modal retrieval \cite{qi2016sketch,bai2018optimization} has been extensively studied in the keyword-based web document retrieval \cite{ensan2017document}, content-based image retrieval \cite{noh2017large},
and product recommendation \cite{kang2017visually,kang2019complete}.
Cross-modal retrieval \cite{wang2017adversarial,feng2014cross,wei2016cross,cao2016deep,wang2015joint,deng2018triplet} emerges as a promising avenue for efficient indexing and searching among large-scale data with different modalities, and is widely used in search engines \cite{buttcher2016information,harman2019information}, E-commerce \cite{ji2017cross,corbiere2017leveraging}, to name a few.
However, these approaches \cite{nurmi2008product,lin2018regional,corbiere2017leveraging,wei2019rpc,wang2016effective} are typically subject to single modal inputs, which makes them hard to apply to many real-world scenarios where multi-modal information exists in both the queries and targets.

\noindent\textbf{WSOD: Weakly Supervised Object Detection.} WSOD \cite{tang2018pcl,2018Generative,2019Cap2Det} reduces its excessive reliance on fine-grained labels by learning from cheaper or freely-available data.
PCL \cite{tang2018pcl} iteratively generates proposal clusters to facilitate the learning of instance classifiers.
Pseudo labels generated from image labels \cite{2018Generative} and unstructured textual descriptions like captions \cite{2019Cap2Det} are also beneficial for boosting the performance of WSOD.
However, WSOD typically relies on a fixed-size collection of predefined classes and is not readily applicable to our proposed task where class labels are not available and categories can be updated dynamically.

\noindent\textbf{Cross-Modal Self-Supervised Learning.}
Existing Vision-language pre-trained models typically use a multi-layer Transformer \cite{vaswani2017attention} architecture such as BERT \cite{devlin2018bert} to learn image-text semantic alignment on multi-modal data.
Single-stream models \cite{li2019visualbert,Su2020VL-BERT:,chen2020uniter} encode the combined multi-modal features in a unified architecture while other two-stream models \cite{lu2019vilbert,tan2019lxmert} instead utilize different encoders for inputs of different modalities.
These methods are not tailored for instance-level retrieval  and we showcase that they might be flawed due to the design defects in the network architecture and the inappropriate pretext tasks.

\section{Instance-Level Retrieval on Product1M}
\subsection{Task Definition}
\label{sec_task_definition}

A product sample $(I, C)$ is an image-text pair where $I$ is the product image and $C$ is the caption.
    Given the \textit{gallery} set of single-product samples $\mathcal{S}=\{\mathcal{S}_i |\mathcal{S}_i= (I_\mathcal{S}^i,C_\mathcal{S}^i)\}$ and the set of multi-product samples $\mathcal{P}=\{\mathcal{P}_i | \mathcal{P}_i=(I_\mathcal{P}^i,C_\mathcal{P}^i)\}$, the task is to retrieve and rank the single-products that appear in the query sample $\mathcal{P}_i$, i.e., to predict a list $RETR^i = [id^i_1, id^i_2, \cdots, id^i_k, \cdots, id^i_N]~~ \forall \mathcal{P}_i \in \mathcal{P}$, where $id^i_k$ corresponds to a specific single-product sample in $\mathcal{S}$.

\subsection{Dataset Statistics}
\label{sec_statistics}
We collect a large number of product samples of 49 brands from E-commerce websites.
These image-text samples are then manually divided into the single-product and multi-product groups according to the corresponding product information.
Product1M is split into the \textit{train}, \textit{val}, \textit{test}, and \textit{gallery} set.
The \textit{train} set contains 1,132,830 samples including both the single-product and multi-product samples, while there are only multi-product samples in the \textit{val} and \textit{test} set, which contain 2,673 and 6,547 samples respectively.
The \textit{gallery} set has 40,033 single-product samples for 458 categories, 392 of which appear in the \textit{val} and \textit{test} set and the remaining ones act as interference items for validating the robustness of a retrieval algorithm.
The samples in the \textit{gallery}, \textit{val}, and \textit{test} sets are annotated with class labels for the purpose of evaluation, i.e., they are not involved in the training process, and the samples in the \textit{train} set are not annotated.
The statistics of Product1M are shown in Table \ref{tab_statistics} and Figure \ref{fig_challenge_statistics}.
More visualizations of Product1M and comparisons with related datasets can be found in the Supplementary Materials.

\begin{figure*}
\setlength{\belowcaptionskip}{-0.6cm}
    \centering
    \includegraphics[width=1.0\textwidth]{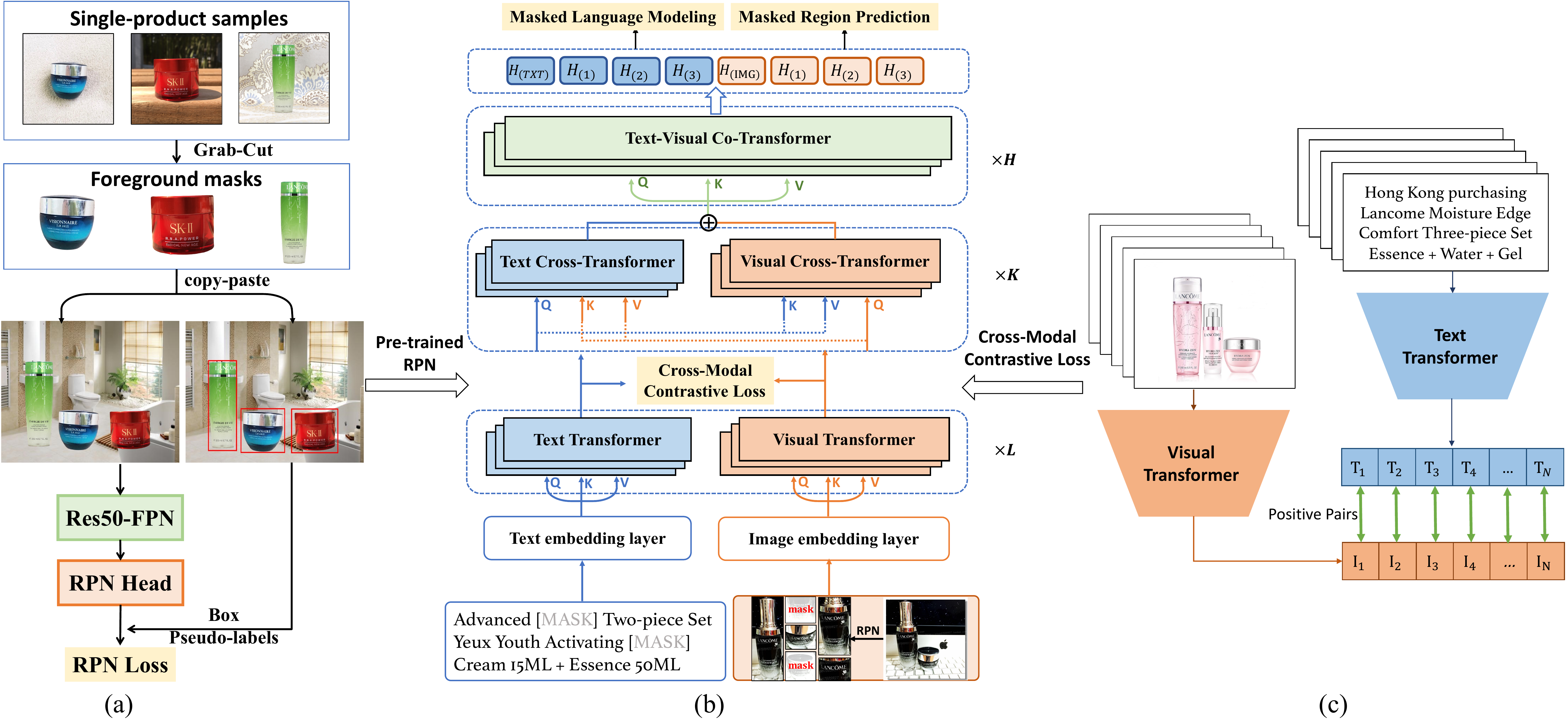}
    \caption{An overview of our instance-level retrieval pipeline. (a) Pretrain an RPN based on pseudo-labels generated by a copy-and-paste data augmentation scheme. (b) Utilize CAPTURE to capture the potential synergy across modality via a hybrid-stream architecture and several pretext tasks. (c) Construct positive pairs of matched image-text samples for cross-modal contrastive learning. Best viewed in color.}
    \label{fig_overall_CAPTURE}
\end{figure*}

\subsection{Dataset Characteristics}
\label{sec_challenge}

\noindent \textbf{Multi-product nature and complex combinations:}
The multi-product images are ubiquitous on E-commerce websites and serve as the query images of instance-level product retrieval.
    As is shown in Figure \ref{fig_challenge_statistics}(1a), products can be organized in abundant forms and layouts and the number of instances can be large.
    The excessive amount and great diversity of fine-grained single-product samples give rise to the complex combinations in different portfolio images.

\noindent \textbf{Weak supervision and fuzzy correspondence:}
    We consider using data of two common modalities, i.e., images and texts, for retrieval.
    Unlike other datasets with clean class labels,
    the supervision from commodity captions is weak and often uninformative.
    We show different types of challenging samples in Figure \ref{fig_challenge_statistics}(1b).
    Some samples contain abbreviations, i.e., a shortened form of several products, in their captions.
    However, the abbreviation like `eight-piece set' does not contain any specific information about products.
    The second type of sample carries irrelevant information, where the commodities described in the title may not appear in the image or vice versa.
    The wide distribution of fuzzy correspondence between images and titles makes it even more challenging for instance-level retrieval.



\noindent \textbf{Consistency with real-world scenarios:}
    We show some challenging samples in Figure \ref{fig_challenge_statistics}(1c).
    They can have a complex background with irrelevant objects, amorphous watermarks, or significant clutter covering the product information.
    Some products of different categories can have almost the same appearance except that the words on the packing are slightly different, e.g., \textit{day cream} vs \textit{night cream}.
    As is shown in Figure \ref{fig_challenge_statistics}(2a,2b), the long-tailed distribution of Product1M aligns well with real-world scenarios.

\section{Methodology}
As is depicted in Figure \ref{fig_overall_CAPTURE}, our framework consists of an augmentation-based detector and a self-supervised multi-modal transformer.
In this section, we first elaborate the training process of RPN and the architectural design of CAPTURE in Section \ref{sec_rpn} and Section \ref{sec_CAPTURE_arch}.
Then we describe two kinds of pretext tasks that enables the self-supervised learning of CAPTURE in Section \ref{sec_capture_masked} and Section \ref{sec_capture_contrastive}.
Finally, we illustrate the inference process for instance-level retrieval in Section \ref{sec_inference}.

\subsection{Training RPN for Multi-Product Detection}
\label{sec_rpn}
Retrieval based simply on the image-level features will lead to an undesirable condition where the retrieval results are overwhelmed by the dominated product in an image.
Thus it is crucial to distinguish different products and extract proposal-wise features in a multi-product image.
While many pre-trained detectors are available, they are infeasible to directly apply to multi-product detection due to the distribution difference between datasets.
Thus we utilize a simple yet effective data augmentation scheme to train a Region Proposal Network (RPN) \cite{ren2015faster} based solely on the single-product images as shown in Figure \ref{fig_overall_CAPTURE}(a).
We first use GrabCut \cite{Meng2014GrabCut} to obtain the foreground masks of single-product images.
With real-world background images from Places365 \cite{zhou2017places}, a copy-and-paste augmentation \cite{dwibedi2017cut} is applied to these foreground masks and background images to generate synthesized images.
In this way, we are able to train a well-performing multi-product detector.
Given the detected regions of RPN, we utilize RoIAlign \cite{he2017mask} to obtain instance-wise features, which are then fed into CAPTURE for further cross-modal learning.
More visualizations and details about the synthesized images and the training of RPN can be found in the Supplementary Materials.

\subsection{Architectural Design of CAPTURE}
\label{sec_CAPTURE_arch}
After training the RPN, we can generate high-quality proposals for different products in an image.
Different from the prevalent single-stream or two-stream transformer architectures, we propose CAPTURE that combines these two architectures into a unified one by stacking three types of layers for semantic alignment and joint learning of multi-modal inputs.
Details are shown in Figure \ref{fig_overall_CAPTURE}(b).
To be specific, the Text/Visual Transformer takes as input the embeddings of the texts or image and is responsible for intra-modal feature learning.
The Text/Visual Cross-Transformer aims to capture and model the inter-modal relations between texts and image by exchanging key-value pairs in the multi-headed attention mechanism.
After that, the features of texts and image are concatenated and serve as the query, key, and value inputs to the Co-Transformer for joint learning of multi-modal features.
These three types of transformer are stacked $L, K$, and $H$ times respectively.
We verify the effectiveness of our architectural design in Table \ref{tab:ablation_layer}.

\begin{table*}[h]
\centering
\resizebox{\textwidth}{!}{
\begin{tabular}{l|ccc|ccc|ccc}
\toprule[1pt]
Method& {\footnotesize{}{mAP@10}} & {\footnotesize{}{mAP@50}} & {\footnotesize{}{mAP@100}} & {\footnotesize{}{mAR@10}} & {\footnotesize{}{mAR@50}} & {\footnotesize{}{mAR@100}} & {\footnotesize{}{Prec@10}} & {\footnotesize{}{Prec@50}} & {\footnotesize{}{Prec@100}} \tabularnewline
\midrule[1pt]

\textit{Image-based} & 40.35  &  36.77  & 34.76   & 17.20  & 15.86  & 15.45  & 32.80  & 30.54  & 29.97  \\

\textit{Text-based} & 61.56 &  59.38  & 58.42   & 23.65  & 22.04  & 20.13  & 56.15  & 57.47  & 57.45  \\
ViLBERT \cite{lu2019vilbert} & 70.11  &  68.19  &  68.29  &  29.05 & 25.54  & 25.02  &  64.64 & 66.35  &  66.60 \\
LXMERT \cite{tan2019lxmert} & 71.37  & 67.83   &  66.73  & 29.83  & 23.15  &  23.89 & 65.97  &  64.79 & 64.77  \\
CLIP* \cite{radford2021learning}  & 70.25  & 69.28   &  67.30  &  29.45  &  25.61  & 25.61 & 67.77  &  68.00 & 68.38 \\
VL-BERT \cite{Su2020VL-BERT:} & 72.01  & 68.22   & 67.79   &  29.15 & 25.59  & 26.16  &  65.25 & 66.92  & 66.64  \\
VisualBERT\cite{li2019visualbert} &  72.27 & 69.60   &  68.28  & 31.69  &  26.31 & 26.83  & 67.31  &  66.48 &  66.62 \\

UNITER \cite{chen2020uniter} & 74.69  & 71.02   &  70.93  &  29.47 & 25.82  &  26.20 & 70.11  &  69.15 & 68.95  \\

\textbf{CAPTURE (Ours)} & \textbf{79.36} & \textbf{74.79} & \textbf{74.63} & \textbf{34.69} &  \textbf{30.04} & \textbf{30.08} & \textbf{73.97} & \textbf{72.12} & \textbf{73.86}   \tabularnewline
\bottomrule[1pt]
\end{tabular}}
\caption{Comparison with different intra- and cross-modal self-supervised baselines.}
\label{tab_performance}
\end{table*}

\subsection{CAPTURE by Masked Multi-Modal Learning}
\label{sec_capture_masked}
We utilize several pretext tasks to enable the self-supervised learning of CAPTURE.
For modality-wise feature learning, we adopt two masked multi-modal modeling tasks, i.e., Masked Language Modeling task (MLM) and Masked Region Prediction task (MRP), following the standard BERT \cite{devlin2018bert} and VisualBERT \cite{li2019visualbert}.
Concretely, for MLM and MRP, approximately 15\% of texts and proposal inputs are masked out and the remaining inputs are used to reconstruct the masked information.
The MLM is handled as in BERT \cite{devlin2018bert}.
For the MRP, the model directly regresses the masked features, which is supervised by the features extracted by the pretrained RPN with a MSELoss.
As for inter-modal relation modeling, Image-Text Matching task (ITM) is widely adopted in many previous methods \cite{li2019visualbert,chen2020uniter,lu2019vilbert,tan2019lxmert}.
Typically, the model is asked to predict whether the text is the corresponding description of an image, which is formulated as a binary classification task.
To generate negative samples, either the image or caption is randomly substituted.
We argue that ITM could be problematic to the fine-grained understanding of an image-text sample at the instance-level.
We hypothesize the deterioration stems from the unmatched image and caption pairs after substitution, which results in the inconsistency between detected regions and text.
We further experimentally validate this claim in Table \ref{tab:loss_ablation}.

\subsection{CAPTURE by Cross-Modal Contrastive Loss}
\label{sec_capture_contrastive}
Aside from intra-modal feature learning, CAPTURE is expected to generate coherent representations of multi-modal inputs and learn the correspondence between them.
To this end, we resort to inter-modality contrastive learning \cite{chen2020simple,radford2021learning} to reach alignment between image and text.
For a minibatch of $N$ image-text samples, there are $2N$ data points in total.
We treat the corresponding image-text pairs as $N$ positive pairs, and the other $2(N-1)$ unmatched pairs are regarded as negative ones.
Formally, given an image-text pair $(x_i,x_j)$ and their encoded features $(\tilde{x}_i, \tilde{x}_j)$, the cross-modal contrastive loss for this positive pair is computed as:
\begin{equation}
   \mathcal{L}(x_i, x_j)=-\log \frac{\exp \left(\operatorname{sim}\left(\tilde{x}_i, \tilde{x}_j\right) / \tau\right)}{\sum_{k=1}^{2 N} \mathbbm{1}_{[k \neq i]} \exp \left(\operatorname{sim}\left(\tilde{x}_i, \tilde{x}_k\right) / \tau\right)},
\end{equation}
where $\operatorname{sim}(\boldsymbol{u}, \boldsymbol{v})=\boldsymbol{u}^{\top} \boldsymbol{v} /\|\boldsymbol{u}\|\|\boldsymbol{v}\|$ computes the cosine similarity of $(\boldsymbol{u}, \boldsymbol{v})$ pairs, $\tau$ denotes the temperature parameter, $\mathbbm{1}_{[k\ne i]}$ is a binary indicator function that returns 1 iff $k \ne i$.
This form of contrastive loss  encourages the encoded features of positive pairs from different modality to be similar while discriminates those of negative ones.
We find it beneficial to inject this supervision at the Text/Visual Transformer and further discussion about the effect of cross-modal contrastive loss can be found in Section \ref{sec_pretext_ablation}.


\subsection{Inference for Instance-Level Retrieval}
\label{sec_inference}
For both the single- and multi-product samples, the proposal-wise features extracted via the pre-trained RPN and the captions are used as input to CAPTURE.
During inference, the Co-Transformer layer outputs $H_{IMG}$ and $H_{TXT}$ as the overall representations of visual and linguistic inputs, respectively.
These two vectors are multiplied together to derive the joint representations of an instance.
Furthermore, since Text/Visual Transformer is supervised with cross-modal contrastive loss, we find it beneficial to concatenate the features of this layer for retrieval.
The resulting features then serve as the input of our retrieval algorithm.
After computing the cosine similarity matrix between an instance and the samples in the $gallery$ set, we retrieve the corresponding single-product samples with the highest similarities for each query.

\section{Experiments}
\label{sec_exp}

\subsection{Implementation Details}
\label{sec_implement}
We attach RPN to a ResNet-50 \cite{he2016deep} backbone pre-trained on ImageNet and follow the training schedule in \cite{ren2015faster}.
We use the BERT \cite{devlin2018bert} to initialize the linguistic transformer of our CAPTURE.
The number of the Text/Visual Transformer, Text/Visual Cross-Transformer, and Co-Transformer is set to $L=4, K=4,$ and $H=4$, respectively, which adds up to 12 transformer layers.
We set the hidden state size of CAPTURE and other baselines to 768 for a fair comparison.
We separately attach a 512-d fully connected layer after Co-Transformer and Text/Visual Transformer for masked multi-modal learning and cross-modal contrastive learning.
The concatenation of the features from these two layers results in a 1024-d feature vector for retrieval, which is also the same for other baselines.
The maximum sequence length for the sentence is set to 36.
We train CAPTURE with a total batch size of 128 for 10 epochs on 4 RTX 2080 GPUs.
We use Adam \cite{kingma2014adam} optimizer with an initial learning rate of 1e-4 and a linear learning rate decay schedule is adopted.
Temperature parameter $\tau$ is set to 0.07.
At inference, CAPTURE takes as input the texts and proposal-wise features to generate instance features.
For a fair comparison with other baselines, we adopt the same training procedure and evaluation protocol in all experiments unless otherwise stated and we use the same augmentation-based RPN for the baselines in Table \ref{tab_performance}.
More details can be found in the Supplementary Materials.

\noindent\textbf{Evaluation Metrics.}
We adopt Precision ($\mathrm{Prec}@N$), mean Average Precision ($\mathrm{mAP} @ N$) and mean Average Recall ($\mathrm{mAR} @ N$) as our evaluation metrics, among which $\mathrm{Prec}@N$ and $\mathrm{mAP} @ N$ are widely used in the retrieval literature \cite{weyand2020GLDv2,CGW17}.
Since exhaustively retrieve every single product is unnecessary and impractical in many scenarios, we report $\mathrm{mAP}$, $\mathrm{mAR}$, and $\mathrm{Prec}$ for $N = 10, 50, 100$.
The details of evaluation metrics can be found in the Supplementary Materials.

\subsection{Weakly-Supervised Instance-Level Retrieval}
\label{sec_main_results}
We compare CAPTURE with several intra- and cross-modal baselines and the results are shown in Table \ref{tab_performance}.

\noindent \textbf{Intra-Modal Schemes.}
We compare our method to two intra-modal schemes including \textit{Image-based} and \textit{Text-based} schemes.
For image-based retrieval, we stack the Visual Transformer layer described in Section \ref{sec_CAPTURE_arch} and adopt the same image input and pretext task, i.e., Masked Region Prediction as CAPTURE.
For text-based retrieval, we stack the Text Transformer layer and use only the text input and Masked Language Modeling pretext task.
We further double the depth of these two models to 24 layers to keep the same amount of parameters as CAPTURE.
It turns out that these two schemes are lagging far behind since they are subject to data of single modality, which suggests that the modeling of the relations between multi-modal data is indispensable.
We provide more experiment results to validate this point in Section \ref{sec_layer_ablation}.


\noindent\textbf{Cross-Modal Schemes.}
We compare CAPTURE to several prevailing self-supervised cross-modal pretraining methods in Table \ref{tab_performance}, including SOTA single-stream and two-stream Vision-language models as well as a SOTA zero-shot classification model, i.e., CLIP \cite{radford2021learning}.
The CLIP* baseline refers to a CLIP-like architecture that uses separate transformers to encode image and text and is trained with a contrastive objective.
Notably, CAPTURE outperforms all these baselines in all three metrics for instance-level retrieval.
Two-stream models, i.e., ViLBERT \cite{lu2019vilbert}, LXMERT \cite{tan2019lxmert} and CLIP*, are generally worse than single-stream ones, which suggests that the fusion mode of multi-modal features is one of the critical factors.
We attribute the superior performance of CAPTURE to its hybrid-stream architecture and we study the impact of different layer types in Section \ref{sec_layer_ablation}.



\begin{table}[]
\label{tab:loss_ablation}
\centering
\footnotesize
\resizebox{\columnwidth}{!}{
\begin{tabular}{c|cccc|c}
\toprule[1pt]
\#&Masked & ITM & CTR & Concat & mAP/mAR/Prec \\

\midrule[1pt]
1& \checkmark&&&     & 72.1 / 28.9 / 72.7 \\
2& \checkmark& & &  \checkmark
 & 71.9 / 28.5 / 72.9 \\
3& \checkmark& \checkmark &&    &  70.2 / 27.1 / 70.2 \\
4&  \checkmark&  & \checkmark &   &  73.3 / 29.1 / 73.2  \\
5& \checkmark&  &\checkmark&\checkmark    & \textbf{74.6} / \textbf{30.1} / \textbf{73.9}\\

\toprule[1pt]
\end{tabular}
}
\caption{The impact of different pretext tasks and cross-modal contrastive loss. Evaluation for $N=100$. `Masked' stands for two masked multi-modal pretext tasks, i.e., MLM and MRP. `CTR' stands for cross-modal contrastive loss.}
\label{tab:loss_ablation}
\end{table}

\subsection{Impact of Pretext Tasks and Contrastive Loss}
\label{sec_pretext_ablation}
As shown in Table \ref{tab:loss_ablation}, ITM will hurt the accuracy of instance-level retrieval (\#1 vs \#3), since it gives rise to mismatch samples, which might be detrimental to the fine-grained understanding of a multi-product image.
We apply the cross-modal contrastive loss at the Text/Visual Transformer layer to align the representations of image and text, which further benefits the learning of consequent layers.
The inclusion of contrastive loss encourages our model to maximizes the feature similarity of positive pairs, which improves all three metrics by 1.2, 0.2, and 0.5, respectively (\#1 vs \#4), and we find it of little help when added to the deeper layers.
Moreover, after concatenating the features from the Text/Visual Transformer with that from the Co-Transformer for retrieval, it further improves all three metrics by 1.3, 1.0, and 0.7, respectively (\#4 vs \#5).
However, we find this concatenation operation will slightly degrade the performance of the model without contrastive loss (\#1 vs \#2), which suggests that the improvement mainly comes from the contrastive loss instead of the operation itself.

\begin{table}[h]
\centering
\footnotesize
\setlength{\belowcaptionskip}{-0.5cm}
\resizebox{\columnwidth}{!}{
\begin{tabular}{lcc|c}
\toprule[1pt]
Model & Config & Depth & mAP/mAR/Prec
\tabularnewline
\midrule[1pt]

w/o-Cross & (6,0,6) & 12 & 73.8 / 28.2 / 71.5 \\
w/o-Co & (6,6,0) & 12 & 73.2 / 29.3 / 72.8 \\
w/o-Txt/Vis & (0,6,6) & 12 &  69.3 / 25.4 / 68.4  \\
\midrule[0.7pt]
CAPTURE-A & (2,5,5) & 12 &  72.8 / 29.0 / 71.3   \\
CAPTURE-B & (5,2,5) & 12 & 73.7 / 29.1 / 71.8  \\
CAPTURE-C & (5,5,2) & 12 &  73.8 / 29.5 / 72.0  \tabularnewline
\midrule[0.7pt]
CAPTURE-S & (2,2,2) & 6 & 67.7 / 25.7 / 68.3 \\
CAPTURE-L & (8,8,8) & 24 &  74.7 / 30.9 / 74.2   \\
\midrule[0.7pt]
\textbf{CAPTURE} & \textbf{(4,4,4)} & \textbf{12} & \textbf{74.6 / 30.1 / 73.9} \\

\bottomrule[1pt]
\end{tabular}}
\caption{Performance of different layer configurations.
Evaluation for $N=100$.}
\label{tab:ablation_layer}
\end{table}

\subsection{Impact of Layer Configuration}
\label{sec_layer_ablation}
We investigate how the configuration of transformer layers will affect the performance of our model in Table \ref{tab:ablation_layer}.
The triplet in the Config column stands for the number of Text/Visual Transformer, Cross-Transformer, and Co-Transformer layer, respectively.
We first remove layers of a specific type while keeping the depth of the resulting network the same as CAPTURE's, i.e., 12 layers, for a fair comparison.
The `w/o-Cross', `w/o-Co', and `w/o-Txt/Vis' refer to the resulting model after removing the Cross-Transformer, Co-Transformer, and Text/Visual Transformer layers from CAPTURE.
As can be seen, the performances of these three models are inferior to that of CAPTURE, which demonstrates the effectiveness of its hybrid-stream architecture.
Moreover, in the second group of Table \ref{tab:ablation_layer} (CAPTURE-A,B,C), we study the combination of three layer types in different proportions.
It turns out that the (4,4,4) configuration achieves the best performance.
We further explore the performance of a smaller model (CAPTURE-S) and a larger model (CAPTURE-L).
As can be seen, CAPTURE with the (4,4,4) configuration achieves a better trade-off between accuracy and parameters.

\subsection{Zero-Shot Instance-Level Retrieval}
\begin{figure}
    \centering
    \includegraphics[width=0.5\textwidth]{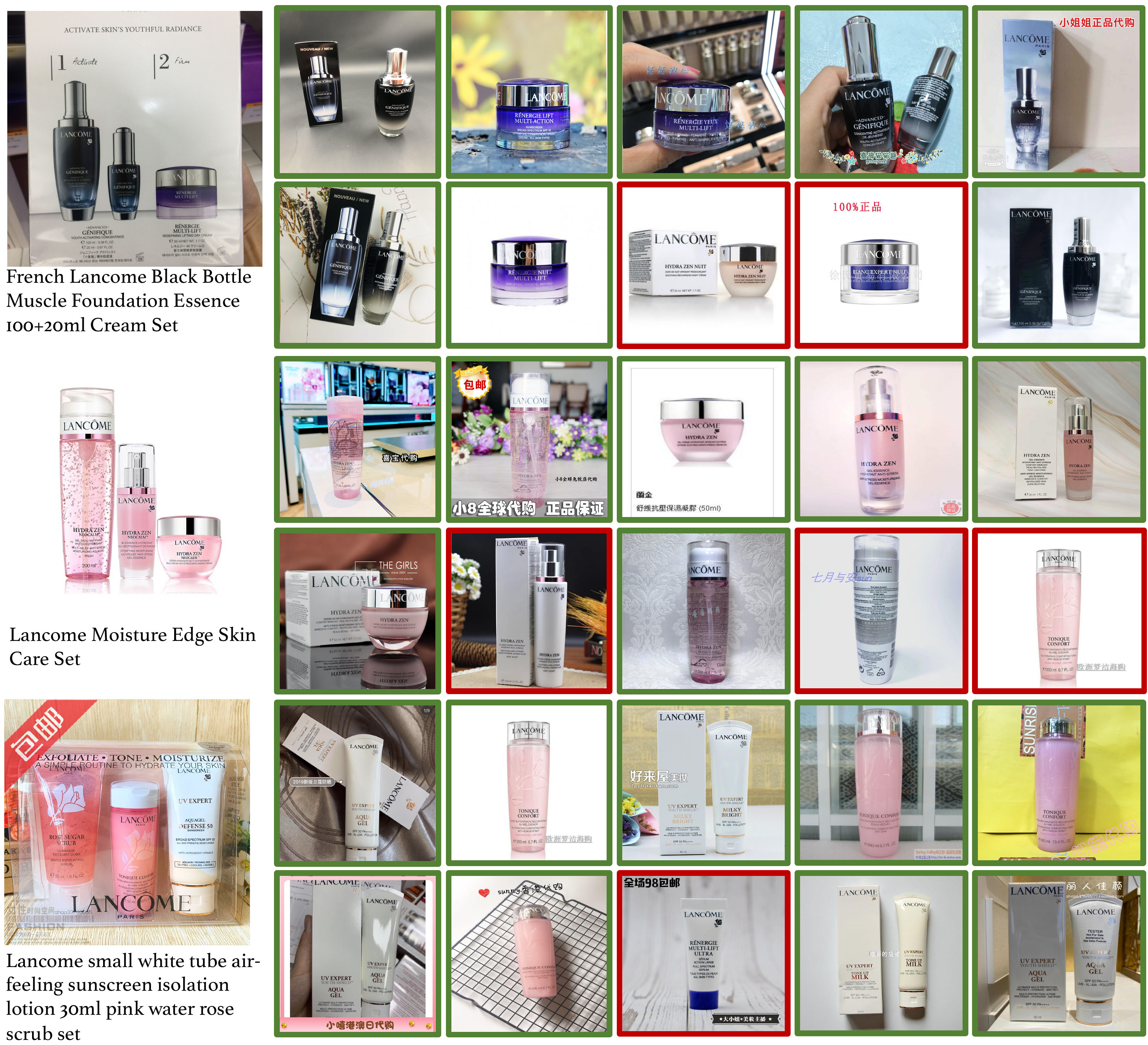}
    \caption{Visualizations of the retrieval results generated by CAPTURE.
    Multi-product query images are on the left.
    Correct/Incorrect retrieval images are highlighted in green/red boxes.
    }
    \label{fig_visualize}
\end{figure}

We argue that a retrieval-based solution generalizes better to real-world scenarios where the category set is updated continuously and large quantities of
 clean labels are too costly to collect.
Unlike detection, our retrieval-based framework does not rely on a fixed-size collection of predefined classes or fine-grained box annotations.
To emphasize this, we conduct zero-shot retrieval experiments and report the results in Table \ref{tab:zero_shot}.
We manually remove 5/10/20 brands from the \textit{train} set and train CAPTURE on the remaining samples so that the removed categories are not disposed to our model during training.
Then we evaluate CAPTURE on the classes of these unseen brands.
We further compare our model with a two-stream model LXMERT and a single-stream model UNITER.
As can be seen, CAPTURE achieves better performance than LXMERT and UNITER for all three metrics, which well demonstrates its generalization capacity.
We also visualize the embeddings generated by CAPTURE and UNITER via t-SNE \cite{van2008visualizing} in Figure \ref{fig_visualize}.
It turns out that the features encoded by CAPTURE are more discriminative, which thus benefits the retrieval task.

\begin{table}
\centering
\setlength{\belowcaptionskip}{-0.3cm}
\footnotesize
\resizebox{\columnwidth}{!}{
\begin{tabular}{c|c|c|c}
\toprule[1pt]
 & Metric & $N=10$ &  $N=50$  \tabularnewline
\midrule[1pt]
\multirow{3}{*}{\rotatebox{90}{\scriptsize 5 brands}} &
mAP$@N$ &  63.3 / 64.5 / \textbf{67.1} &  60.7 / 62.2 / \textbf{64.4}  \tabularnewline
& mAR$@N$ & 23.2 / 24.7 / \textbf{25.9}  &  19.2 / 20.1  / \textbf{20.9}   \tabularnewline
&Prec$@N$ & 56.5 / 57.1 / \textbf{60.5}  &  56.5 / 57.7 / \textbf{62.4}   \tabularnewline
\midrule[0.7pt]
\multirow{3}{*}{\rotatebox{90}{\scriptsize 10 brands}} &
mAP$@N$ & 56.8 / 58.2 / \textbf{61.5} &  54.1 / 55.2 / \textbf{57.1}   \tabularnewline
&mAR$@N$ & 19.6 / 20.5 / \textbf{24.0}  &  16.4 /17.4 / \textbf{18.4}   \tabularnewline
&Prec$@N$ & 50.2 / 51.9 / \textbf{53.0}  &  51.3 / 52.7 / \textbf{54.3}   \tabularnewline
\midrule[0.7pt]
\multirow{3}{*}{\rotatebox{90}{\scriptsize 20 brands}}
& mAP$@N$ & 42.6 / 43.3 / \textbf{44.4}  & 36.7 / 37.3 / \textbf{38.8}   \tabularnewline
&mAR$@N$ & 17.5 / 17.6 / \textbf{17.9}  &  12.9 / 13.2 / \textbf{13.5}   \tabularnewline
&Prec$@N$ & 32.2 / 32.4 / \textbf{34.0}  &  32.9 / 33.1 / \textbf{34.5}  \tabularnewline
\bottomrule[1pt]
\end{tabular}}
\caption{Performance comparison of zero-shot retrieval. Organized in the order of LXMERT/UNITER/CAPTURE.}
\label{tab:zero_shot}
\end{table}





\begin{table}
\centering
\setlength{\belowcaptionskip}{-0.3cm}
\footnotesize
\resizebox{\columnwidth}{!}{
\begin{tabular}{l|ccc}
\toprule[1pt]
Method& {\footnotesize{}{mAP$@100$}} &  {\footnotesize{}{mAR$@100$}} &  {\footnotesize{}{Prec$@100$}} \tabularnewline
\midrule[1pt]
UNITER-single & 86.56 & 80.82 & 80.82 \tabularnewline
LXMERT-single & 86.05 & 80.59 & 80.59 \tabularnewline
CAPTURE-single & \textbf{88.24} & \textbf{83.33} & \textbf{83.33} \tabularnewline
\midrule[0.7pt]
CAPTURE-natural & 70.36 & 26.46 & 66.53 \tabularnewline
CAPTURE-1Inst  &  60.03 &   20.43 &  58.42  \tabularnewline
CAPTURE & \textbf{74.63} & \textbf{30.08} & \textbf{73.86}    \tabularnewline
\midrule[0.7pt]
CAPTURE-subset & 73.36  & 30.44 & 72.41 \tabularnewline
CAPTURE-gt & \textbf{77.79} & \textbf{37.40} & \textbf{77.13} \tabularnewline

\bottomrule[1pt]
\end{tabular}}
\caption{Ablation study of single-product retrieval and the impact of detection performance on retrieval. Note that for single-product retrieval, the metric Prec$@N$ is equivalent to mAR$@N$ since there is only one category in an image.}
\label{tab:ablation_detector}
\end{table}

\subsection{Comparisons on Single-Product Retrieval}
\label{sec_single_product_ablation}
It is noteworthy that CAPTURE is applicable to both single-product  and multi-product retrieval.
Indeed, it excels on these two tasks and achieves better performance than other baselines in single-product retrieval.
Specifically, for each single-product sample in the \textit{gallery} set, we pick it out as a query and perform single-product retrieval among the remaining samples in the \textit{gallery} set.
We compare the performance of three models, i.e., UNITER-single, LXMERT-single and CAPTURE-single, in Table \ref{tab:ablation_detector}.
As can be seen, the performance of single-product retrieval is much higher than that of multi-product retrieval since the difficulty is largely reduced when there is only one instance/entity in the image/text.
Furthermore, we notice the performance of `CAPTURE-single' is still better than that of two other baselines, which further demonstrates the superiority of CAPTURE.

\begin{figure}
    \centering
    \includegraphics[width=0.5\textwidth]{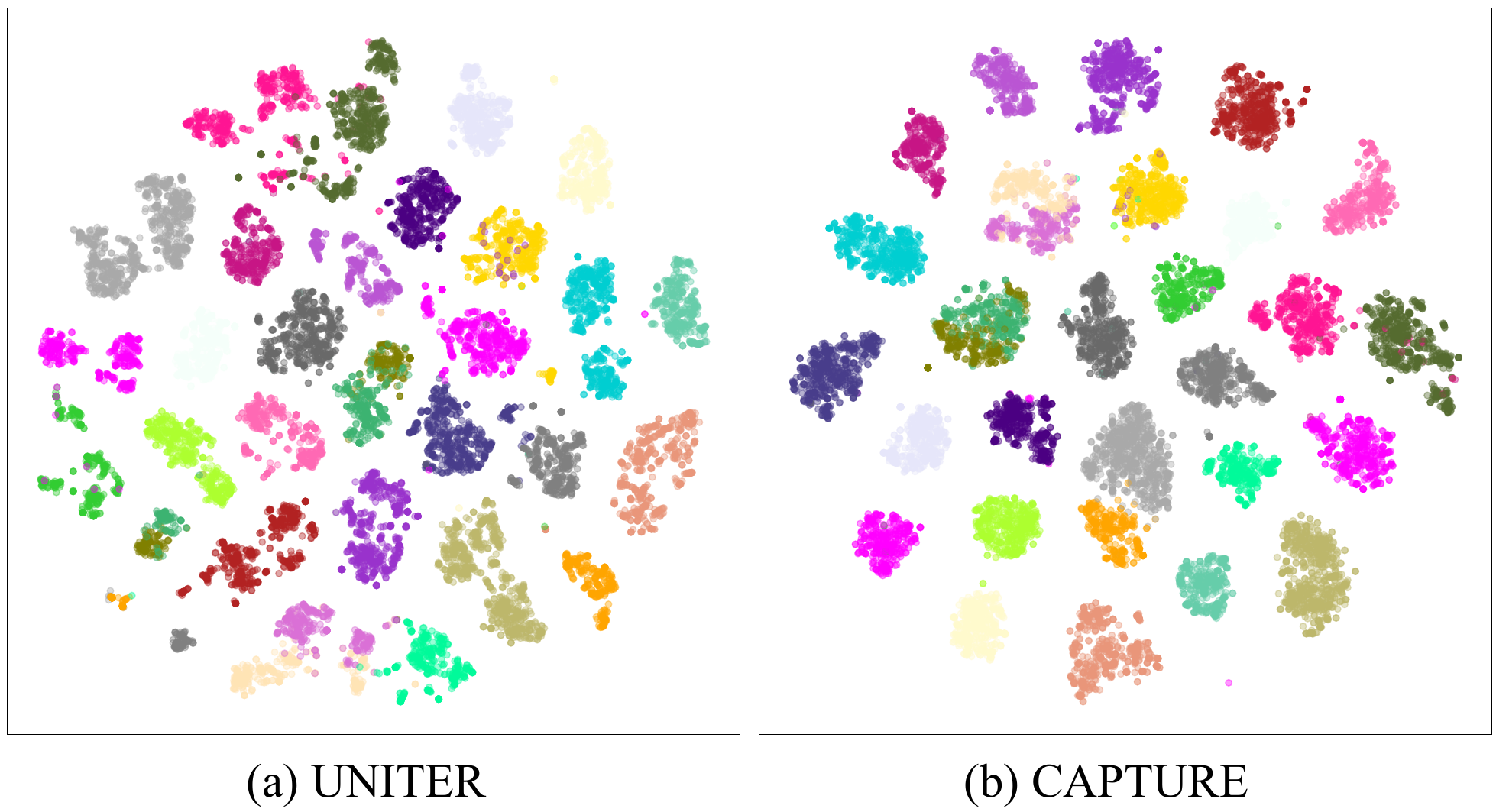}
    \caption{Visualize the embeddings generated by CAPTURE and UNITER via t-SNE.
    Points belonging to the same category are of the same color. Best viewed in color.}
    \label{fig_embedding}
\end{figure}

\subsection{Impact of Detection Performance on Retrieval}
\label{sec_detector_ablation}
We conduct several experiments to explore how the performance of a detector will influence instance-level retrieval.
The results are listed in Table \ref{tab:ablation_detector}.
As we claim in Section \ref{sec_rpn}, the off-the-shelf pretrained detectors are not readily applicable to our dataset due to the distribution difference between natural images and commodity images.
To verify this, we replace the RPN with Faster R-CNN \cite{ren2015faster} pre-trained on Visual Genome \cite{krishna2017visual} and utilize it to generate instance-wise input features of CAPTURE.
The resulting model, named `CAPTURE-natural', is inferior to CAPTURE in all three metrics.
For the `CAPTURE-1Inst' model, we feed the whole image and an image-level bounding box, which is of the same size as the image, to CAPTURE for inference.
This scheme performs unsatisfactorily due to the failure of instance recognition, which suggests that the detector may become a performance bottleneck.
Going a step further, to explore the upper bound of CAPTURE, we randomly select 1,338 multi-product images and manually label the bounding boxes of these images.
For the `CAPTURE-subset' model, we simply evaluate CAPTURE on this annotated subset.
For the `CAPTURE-gt' model, the ground truth boxes and their corresponding features serve as the input to CAPTURE.
As can be seen, the performance gap of these two models suggests that the performance of a detector can play an essential role in instance-level retrieval.
Moreover, the mAR gap between them is relatively large, which indicates that the false negatives in detection will hurt the performance of instance-level retrieval.

\section{Conclusion}
In this paper, we present the first effort on extending canonical intra-/cross-modal retrieval to a more generalized setting, i.e., weakly-supervised multi-modal instance-level product retrieval, which has wide application potential in the E-commerce industry.
We contribute Product1M, which is one of the largest multi-modal retrieval datasets as well as the first one specifically tailored for instance-level retrieval.
Aside from that, we propose a novel hybrid-stream transformer, named CAPTURE, that excels in capturing the potential synergy between data of different modalities.
Moreover, we overcome the unmatched issue incurred by the inappropriate pretext task by enforcing cross-modal contrastive learning between multi-modal features.
Extensive experiments demonstrate that our CAPTURE surpasses the SOTA cross-modal pretraining models in terms of all metrics by a large margin.
We hope the proposed Product1M, CAPTURE, and solid baselines will spur further investigation into a more reliable and flexible retrieval system.
\section{Acknowledgement}
This work was supported in part by National Key R$\&$D Program of China under Grant No.2018AAA0100300, National Natural Science Foundation of China (NSFC) under Grant No.U19A2073 and No.61976233, Guangdong Province Basic and Applied Basic Research (Regional Joint Fund-Key) Grant No.2019B1515120039, Guangdong Outstanding Youth Fund (Grant No.2021B1515020061), Shenzhen Fundamental Research Program (Project No.RCYX20200714114642083, No.JCYJ20190807154211365).
{\small
\bibliographystyle{ieee_fullname}
\bibliography{egbib}
}
\end{document}